\pdfoutput=1
\documentclass[letterpaper]{article} % DO NOT CHANGE THIS

\usepackage{aaai25}  % DO NOT CHANGE THIS
\usepackage{times}  % DO NOT CHANGE THIS
\usepackage{helvet}  % DO NOT CHANGE THIS
\usepackage{courier}  % DO NOT CHANGE THIS
\usepackage[hyphens]{url}  % DO NOT CHANGE THIS
\usepackage{graphicx} % DO NOT CHANGE THIS
\usepackage{natbib}  % DO NOT CHANGE THIS AND DO NOT ADD ANY OPTIONS TO IT
\usepackage{caption} % DO NOT CHANGE THIS AND DO NOT ADD ANY OPTIONS TO IT
\usepackage{amsmath}
\usepackage{bm}
\usepackage{multirow}
\usepackage{booktabs}
\usepackage{makecell}
\usepackage[usestackEOL]{stackengine}
\usepackage{algorithm}
\usepackage{algorithmic}

\usepackage{newfloat}
\usepackage{listings}
\DeclareCaptionStyle{ruled}{labelfont=normalfont,labelsep=colon,strut=off} % DO NOT CHANGE THIS
\floatstyle{ruled}
\newfloat{listing}{tb}{lst}{}
\floatname{listing}{Listing}
\title{EditBoard: Towards a Comprehensive Evaluation Benchmark for Text-Based Video Editing Models}

\author{
    Yupeng Chen\textsuperscript{\rm 1}\thanks{Work done when Yupeng was a visiting student at University of Oxford.},
    Penglin Chen\textsuperscript{\rm 2}\equalcontrib,
    Xiaoyu Zhang\textsuperscript{\rm 1}\equalcontrib,
    Yixian Huang\textsuperscript{\rm 1},
    Qian Xie\textsuperscript{\rm 3}\thanks{Corresponding author.}
}
\affiliations{
    \textsuperscript{\rm 1}The Chinese University of Hong Kong, Shenzhen\\
    \textsuperscript{\rm 2}Nanjing University\\
    \textsuperscript{\rm 3}University of Leeds\\
    \{yupengchen, xiaoyuzhang2, yixianhuang1\}@link.cuhk.edu.cn, penglinchen@smail.nju.edu.cn, Q.Xie2@leeds.ac.uk
}

\begin{document}

\maketitle

\begin{abstract}
The rapid development of diffusion models has significantly advanced AI-generated content (AIGC), particularly in Text-to-Image (T2I) and Text-to-Video (T2V) generation. Text-based video editing, leveraging these generative capabilities, has emerged as a promising field, enabling precise modifications to video content based on textual prompts. Despite the proliferation of innovative video editing models, there is a conspicuous lack of comprehensive evaluation frameworks that holistically assess these models’ performance across various dimensions. Existing metrics are limited, inconsistent, and focused on assigning a single score per metric, failing to reveal model's performance on each editing task. To address this gap, we propose EditBoard, the first comprehensive evaluation benchmark for text-based video editing models. EditBoard encompasses nine automatic metrics across four key dimensions, evaluating models on four categories of tasks, and introduces three new metrics to assess fidelity. This task-oriented framework facilitates objective evaluation by breaking down model performance into details, providing insights into each model’s strengths and weaknesses. By open-sourcing EditBoard, we aim to standardize evaluation and advance the development of robust video editing models. 
\end{abstract}

\begin{links}
    \link{Code}{https://github.com/Samchen2003/EditBoard}
\end{links}

\section{Introduction}
Recent years have witnessed the rapid development of diffusion models \cite{sohl2015deep, ho2020denoising}, which have been widely applied in the context of AI-generated content (AIGC), such as Text-to-Image (T2I) generation \cite{nichol2022glide, rombach2022high, li2019controllable, guo2023animatediff} and Text-to-Video (T2V) generation \cite{chen2023videocrafter1,luo2023videofusion, villegas2022phenaki}. Harnessing the generative capabilities of these models, text-based video editing is an emerging field that aims to edit specific parts of the video based on text prompts. 

With the growth of innovative video editing models \cite{wu2023tune, qi2023fatezero, jeong2023ground, geyer2023tokenflow}, there remains a notable lack of comprehensive evaluation benchmarks that holistically assess these models' performance across various dimensions. The automatic metrics currently employed are limited in number and scope. For example, models like FateZero \cite{qi2023fatezero} and Ground-A-Video \cite{jeong2023ground} use only two automatic metrics, focusing on temporal consistency between edited frames and frame-wise editing success rate. Moreover, inconsistent naming conventions across papers hinder unified testing and comparison. Most importantly, current evaluations overlook the diversity of editing tasks and use scores from limited dimensions to represent overall performance.

\begin{figure*}[t]
\centering
\includegraphics[width=1.9\columnwidth]{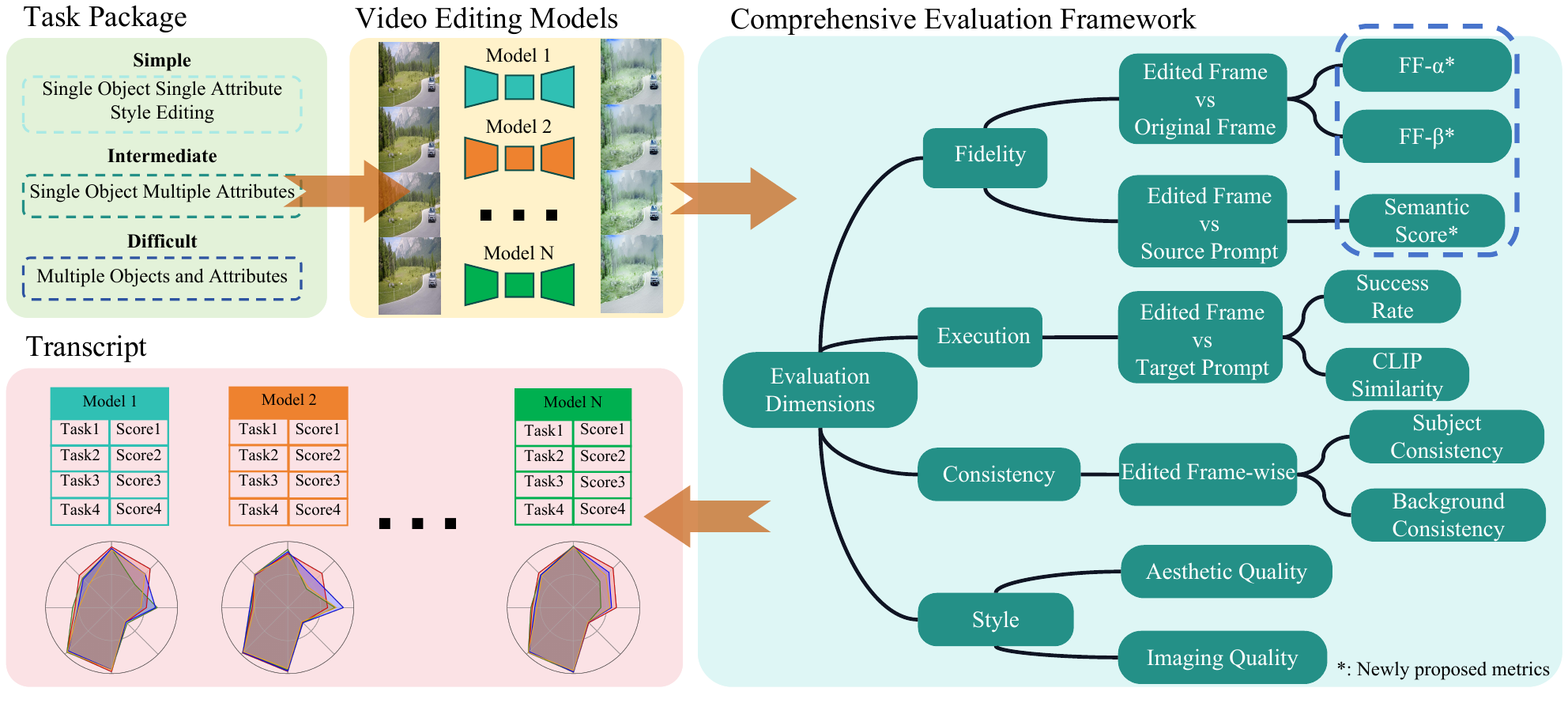} 
\caption{ \textbf{Overview of EditBoard.} We propose EditBoard, the first comprehensive evaluation benchmark for text-based video editing models. We design a task-oriented evaluation benchmark with four dimensions that break down models' performance across multiple levels, facilitating objective evaluation and offering valuable insights. Additionally, we introduce three new metrics and apply nine metrics in total that cover all the evaluation dimensions. EditBoard produces a transcript for each model to discover its advantages and limitations. We also conduct Human Preference Annotation for the edited videos, demonstrating that EditBoard evaluation results align closely with human perception.}
\label{overview}
\end{figure*}

To address these gaps, we propose EditBoard (see Figure~\ref{overview}), the first comprehensive evaluation benchmark for text-based video editing models. EditBoard encompasses nine metrics across four dimensions. First, given the original video, source prompt, and target prompt, we evaluate edited video across three dimensions: \textbf{fidelity} between (1) edited frames and original frames, (2) edited frames and unedited parts of source prompt, \textbf{execution} of target prompt, and \textbf{consistency} between edited frames. For fidelity, we propose FF-$\alpha$ (Frame Fidelity) and FF-$\beta$ to measure motion and structural similarity between edited and original frames, as well as a Semantic Score to assess the accuracy of object-aware editing (the ability to identify the object to be edited and leave other parts unchanged).  For execution, we use Success Rate and CLIP Similarity \cite{hessel2021clipscore, parmar2023zero} to evaluate how well the edited frames match the target prompt. For consistency, we use Subject Consistency and Background Consistency to evaluate whether the frames remain coherent throughout the video. 
Furthermore, we focus on the dimension of \textbf{style} and assess whether the edited video is visually appealing using Aesthetic Quality and Imaging Quality, following the naming conventions from VBench \cite{huang2024vbench}, a benchmark for evaluating video generative models. Additionally, we utilize EditBoard to evaluate five state-of-the-art video editing models, deriving valuable insights from the results. This evaluation highlights each model's strengths and weaknesses, offering possible explanations for their performance. These findings not only enhance our understanding of current models but also propose potential directions for future research.

We notice that a concurrent survey on diffusion model-based video editing \cite{sun2024diffusion} proposes V2VBench, which incorporates existing metrics primarily designed for evaluating video generation. However, these metrics predominantly fall into the dimensions of consistency and execution, leaving significant gaps in fidelity. Our work distinguishes itself by introducing three new automatic metrics and offering a comprehensive evaluation benchmark tailored specifically for video editing models.

Our key contributions can be summarized as follows:
\begin{itemize}
\item We propose the first comprehensive evaluation benchmark for video editing that focuses on four dimensions, having nine metrics in total. We will open-source EditBoard for researchers to thoroughly assess their models.
\item We propose three new metrics to evaluate fidelity between edited videos and original videos/prompts, which align closely with human perception.
\item We define four main tasks of text-based video editing, categorized into simple, intermediate, and difficult levels, enabling a thorough evaluation of models.
% Our experiment provides insights for future research.
\end{itemize}

\section{Related Works}
\subsection{Video Editing}
The development of video generative models \cite{blattmann2023align, chen2023videocrafter1, gupta2023photorealistic, he2022latent, ge2023preserve} has paved the way for advancements in video editing. Unlike video generation, video editing is a more nuanced task that not only involves creation, such as turning a man into Batman, but also requires an understanding of the original structure and adherence to the source's framework. Numerous advanced video editing models have emerged, achieving impressive results through various methods. For example, FateZero \cite{qi2023fatezero}, TokenFlow \cite{geyer2023tokenflow}, and Video-P2P \cite{liu2024video} utilize attention feature injection. Control-A-Video \cite{chen2023control} and VideoControlNet \cite{hu2023videocontrolnet} employ latent manipulation. StableVideo \cite{chai2023stablevideo} and DiffusionAtlas \cite{chang2023diffusionatlas} leverage diffusion atlases. The rapid proliferation of video editing models underscores the need for a comprehensive evaluation benchmark that highlights each model's strengths and weaknesses and provides actionable insights. EditBoard addresses this need by offering the first evaluation benchmark for video editing, defining four evaluation dimensions with nine automatic metrics and four editing tasks.

\subsection{Evaluation of Video Editing Models}

Currently, the major automatic evaluation metrics are summarized as:
\begin{itemize}
\item \textbf{Temporal Consistency (Tem-Con)}: Used in FateZero \cite{qi2023fatezero}, this metric measures temporal consistency by computing the cosine similarity between all pairs of consecutive frames. It is also referred to as Frame-Con in Ground-A-Video \cite{jeong2023ground}, CLIP-F in EVA \cite{yang2024eva} and CLIP-Image in AnyV2V \cite{ku2024anyv2vtuningfreeframeworkvideotovideo}.
\item \textbf{LPIPS}: Utilized by StableVideo \cite{chai2023stablevideo} and VideoControlNet \cite{hu2023videocontrolnet}, this metric is adapted from LPIPS \cite{zhang2018unreasonable}, with LPIPS-P measuring deviation from the original video frames and LPIPS-T measuring deviation between adjacent frames.
\item \textbf{CLIP Score}:  Employed by TokenFlow \cite{geyer2023tokenflow} and StableVideo \cite{chai2023stablevideo}, this metric measures the average similarity between the CLIP embedding of each edited frame and the target text prompt. It is also known as CLIP-T in EVA \cite{yang2024eva}, CLIPSIM in VideoControlNet \cite{hu2023videocontrolnet}, and CLIP-Text in AnyV2V \cite{ku2024anyv2vtuningfreeframeworkvideotovideo}. FateZero \cite{qi2023fatezero} and EVA \cite{yang2024eva} use the percentage of frames where the edited image has a higher CLIP similarity to the target prompt than the source prompt, denoted as Frame Accuracy. 

\item \textbf{Warp Error (Warp-err)}: Used in TokenFlow \cite{geyer2023tokenflow} and EVA \cite{yang2024eva}, this metric computes the optical flow of the original video, warps the edited frames accordingly, and measures the warping error. It is also referred to as Optical Flow Error in VideoControlNet \cite{hu2023videocontrolnet}.
\end{itemize}

Several drawbacks in current evaluation practices are evident. Firstly, metric names are not standardized. Secondly, each model uses only a limited set of automatic metrics. Thirdly, aside from the source video and prompts, editing models are also generative models. However, few of them are evaluated using metrics for generative models. In contrast, V2VBench \cite{sun2024diffusion} primarily employs metrics for generative models, neglecting the need for specifically designed video editing evaluation metrics.
In terms of testing, most models are tested on a limited range of tasks and assigned a single score, failing to reveal their performance on individual tasks. Some models may excel in complex tasks but underperform in simpler tasks compared to baseline models.
To address these gaps, we propose a unified evaluation benchmark. EditBoard focuses on tailored metrics for editing models, supplemented by metrics used in evaluating generative models. Additionally, task-oriented testing breaks down each model's performance into various aspects for thorough evaluation.

\section{Comprehensive Evaluation System}

\subsection{Overview}

We mathematically formulate the problem of text-based video editing as follows: given a sequence of original frames $(f_0, f_1,\ldots, f_n)$ and a source prompt $p_s$ which describes the original video, a model $E$ serves as a function that maps each frame $f_i$ to a new frame $f_i'$ according to the target prompt $p_t$, thus obtaining the edited sequence of frames $(f_0', f_1',\ldots, f_n')$. This process can be expressed as:
\begin{equation}
    E(f_0,f_1,\ldots,f_n; p_s,p_t) = (f_0', f_1',\ldots, f_n')
\end{equation}

When evaluating a video editing model, it is crucial to assess how well it preserves the original video's motion and structure. The first dimension, i.e., fidelity, examines the alignment between edited frames and original frames $(f_0', f_0), (f_1', f_1),\ldots, (f_n', f_n)$, as well as between edited frames and source prompt $(f_0', p_s), (f_1', p_s),\ldots, (f_n', p_s)$. Additionally, the primary emphasis lies in evaluating models' proficiency in performing editing tasks based on target prompts. The second dimension, execution, assesses the models' capability to successfully execute target prompts. Furthermore, the quality of the edited video itself must also be considered. The third evaluation dimension, consistency, addresses the coherence between consecutive frames $(f_0', f_1'), (f_1', f_2'),\ldots, (f_{n-1}', f_n')$. Finally, the fourth dimension, style, focuses on artistic quality and evaluates whether the edited video is visually appealing.

In Subsection \textbf{Evaluation Dimensions}, we detail the four evaluation dimensions, corresponding metrics, and their respective functions. In Subsection \textbf{Task-Oriented Testing}, we elaborate on the four tasks defined for task-oriented testing. In Subsection \textbf{Human Alignment}, we describe the experiments conducted to ensure the alignment of automatic metrics with human perception.

\subsection{Evaluation Dimensions}

EditBoard includes nine metrics covering four dimensions to comprehensively evaluate video editing models.

\subsubsection{Fidelity} The primary focus lies in fidelity, which refers to how accurately the edited frames preserve the motion, structure, and other visual characteristics of the original frames. Unlike video generation, video editing has a mold to follow, which is the original video, making fidelity paramount. This evaluation dimension reveals a model’s ability to comprehend and replicate the patterns of the original video while making the required modifications.

\begin{figure}[t]
\centering
\includegraphics[width=\columnwidth]{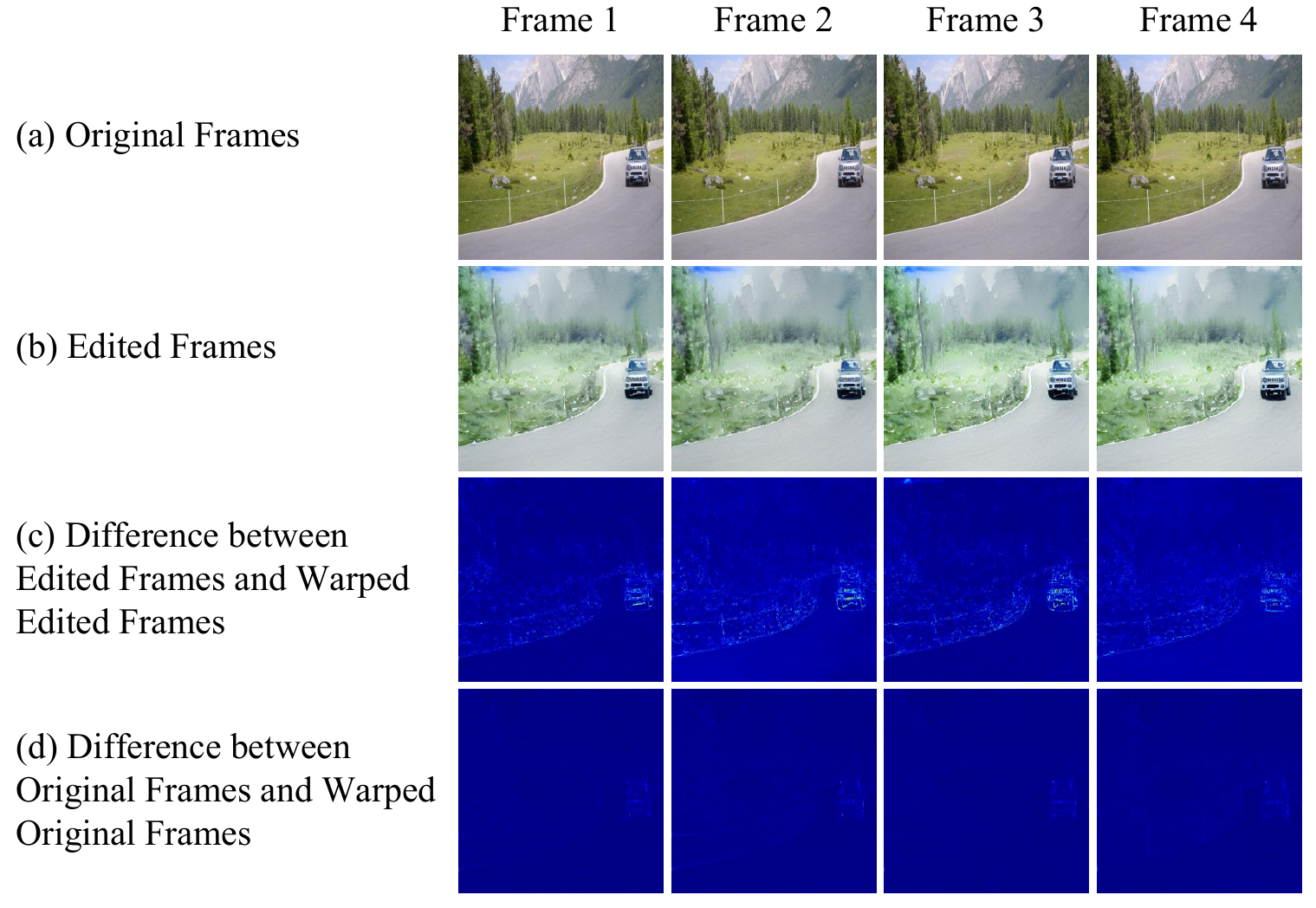} % Reduce the figure size so that it is slightly narrower than the column. Don't use precise values for figure width.This setup will avoid overfull boxes.
\caption{We visualize the errors in reconstructing edited frames and original frames both using optical flows from the original video. For videos satisfying the requirement of FF-$\alpha$, reconstructing original frames yields minor errors compared to reconstructing edited frames. We use the reconstruction error of edited frames to calculate FF-$\alpha$.}
\label{fig2}
\end{figure}

\noindent \textbf{Fidelity - FF-}\bm{$\alpha$}\textbf{.}
  To evaluate motion and structural similarity to the original video and reflect temporal flickering, we propose the FF-$\alpha$ metric. Given the original frames $(f_0, f_1, \ldots, f_n)$ and the edited frames $(f_0', f_1', \ldots, f_n')$, we compute an average score $\frac{1}{n} \sum_{i=0}^{n-1}F(f_i, f_i')$. The original video serves as the ground truth, which typically does not exhibit flickering issues. Using optical flow estimators like PWC-Net and FlowNet \cite{sun2018pwc, ilg2017flownet}, we calculate the optical flows ($l_1, l_2, \ldots, l_n$) from original frame pairs $(f_{i-1}, f_i)$. Let \(\omega:\mathcal{F}\times \mathcal{L}\to \mathcal{F}\) be the WARP function, and denote \(\omega(f_{i+1}, l_{i+1})\) as \(w_i\). The optical flow is then used for backward warping to reconstruct the original frames, resulting in warped frames $(w_0, w_1, \ldots, w_{n-1})$. We also reconstruct the edited frames, obtaining $(w_0', w_1', \ldots, w_{n-1}')$ where

\begin{equation}
    w_i = \omega(f_{i+1}, l_{i+1})
\end{equation}
\begin{equation}
    w_i' = \omega(f_{i+1}', l_{i+1})
\end{equation}
For each pair of reconstructed original frames and original frames, we calculate the absolute difference across the RGB channels and generate a mask $M_i$. If the maximum difference across the three channels is smaller than the threshold $\theta$, the pixel value is set to 1 in the mask; otherwise, it is set to 0. We then calculate the absolute difference between $w_i'$ and $f_i'$ in the valid areas indicated by the mask. The pixel-level score is defined as the maximum difference across the three channels. The score for each frame is then obtained by averaging the pixel-level scores over the valid area ($M_i = 1$):
\begin{equation}
   P(w_i', f_i', M_i) = \frac{1}{|\delta (M_i = 1)|} M_i \cdot ||w_i' -  f_i'|| 
\end{equation}
where $|\delta (M_i = 1)|$ denotes the size of valid area. Finally, we take the average frame score as the FF-$\alpha$ score. Despite the promising results observed during testing with FF-$\alpha$, the reconstruction of frames based on warping demonstrates suboptimal performance when objects undergo significant positional changes, particularly at high sampling rates. When the percentage of valid pixels is low, the evaluation only covers a small portion of the frame. To address this issue, we propose FF-$\beta$ with another threshold $\sigma$. When the valid pixel percentage falls below $\sigma$, FF-$\beta$ is used for evaluation. For those above the threshold, FF-$\alpha$ remains suitable for evaluation (see Figure~\ref{fig2}).
\par

\noindent \textbf{Fidelity - FF-}\bm{$\beta$}\textbf{.} 
FF-$\beta$ takes as input the original frames $(f_0, f_1,\ldots, f_n)$ and the edited frames $(f_0', f_1',\ldots, f_n')$, and outputs the average score $\frac{1}{n} \sum_{i=1}^{n}F'(f_i, f_i')$. 
Instead of using warping and calculating differences across valid pixels, we directly estimate optical flows for original and edited frames, obtaining $(l_1, l_2,\ldots, l_n)$, $(l_1', l_2',\ldots, l_n')$. For better fidelity, we aim for the smallest possible angle between the pairwise flows. Previous methods solely focus on the distance between the endpoints of two optical flows, neglecting the angle, which we consider crucial for motion alignment. Therefore, we use cosine similarity to formulate the pixel-level score as $1-cos\theta$ (where $\theta$ is the angle between the optical flows), aligning it with FF-$\alpha$ such that a lower score indicates better fidelity. Then we compute the average score across the whole frame. The final FF-$\beta$ is given by the averaged frame-level score. 
The rationale for not using FF-$\beta$ directly is that during testing, the score difference is minimal compared to FF-$\alpha$ for videos with small movements between consecutive frames. This can be attributed to the higher sensitivity of pixel-level intensity errors compared to flow errors. Thus, FF-$\alpha$ amplifies performance differences for better comparison when movements are minor.
Additionally, by using warped frames as ground truth to calculate differences, we can simply edit the first frame with state-of-the-art image editing models and employ warping to generate the subsequent frames, eliminating the need for a time and space-consuming video editing model. However, for videos with significant object movement, the warping method fails, whereas advanced video editing models succeed. Therefore, FF-$\beta$ is necessary for such evaluations. In summary, FF-$\alpha$ and FF-$\beta$ are complementary and both are crucial for comprehensive evaluation.

\noindent \textbf{Fidelity - Semantic Score}\textbf{.}
This evaluation dimension, often overlooked, assesses how well the should-be unedited part of the frame remains unedited. Ideally, except for the edited region, the rest of the frame should remain identical to the original, demonstrating accurate or object-aware editing. For this evaluation, we ensure that the target prompt focuses on the object indicated by the semantic mask. The input includes original frames $(f_0, f_1,\ldots, f_n)$, edited frames $(f_0', f_1',\ldots, f_n')$, and semantic masks $(M_0, M_1,\ldots, M_n)$. The output is the average score $\frac{1}{n+1} \sum_{i=0}^{n}S(f_i, f_i', M_i)$. We calculate the difference between the edited and original frames over unmasked regions ($M_i = 0$) by taking the maximum absolute difference across the RGB channels. The process is mathematically formulated as follows:
\begin{equation}
    S(f_i, f_i', M_i) = \frac{1}{|\delta (M_i = 0)|}\bar{M_i} \cdot ||f_i - f_i'|| 
\end{equation}

\subsubsection{Execution} The second evaluation dimension assesses how effectively the edited frames align with the target prompt. In video editing, the goal is to modify the original video to closely match the target prompt while preserving the original video’s structure. While fidelity measures the preservation of the original video’s patterns, execution evaluates the model’s ability to successfully implement the changes required by the target prompt.

\noindent \textbf{Execution - Success Rate.}
Success Rate is a metric that quantifies the effectiveness of frame edits, considering both the source and target prompts. Given the input of edited frames ($f'_0, f'_1,\ldots, f'_n $),
source prompt $p_s$, and target prompt $p_t$, the output is calculated as  $\frac{1}{n+1}\sum_{i=0}^{n} \rho(f'_i, p_s, p_t)$, where $\rho$ is a boolean function, with 1 indicating successful edit and 0 indicating failure. This metric quantifies the percentage of frames where the cosine similarity between the edited frame and the target prompt exceeds the similarity between the edited frame and the source prompt. Each frame is evaluated against both prompts using a pre-trained CLIP model \cite{radford2021learning}. We have revised the name of the metric from Frame Acc, as defined in FateZero, to Success Rate, as this term more accurately encapsulates the function of this metric, which is to assess the percentage of successful editing executions.

\noindent \textbf{Execution - CLIP Similarity.}
CLIP Similarity measures the textual alignment between the edited frames and target prompt $p_t$. Given the input of edited frames $ (f'_0, f'_1,\ldots, f'_n) $ and target prompt \( p_t \), the output is the average CLIP Similarity, calculated as $\frac{1}{n+1}\sum_{i=0}^{n} \text{CLIP}(f'_i, p_t)$. This metric represents the average cosine similarity between the CLIP embeddings of the edited frames and the target prompt. Each frame is encoded into the CLIP feature space and compared against the encoded prompt to generate a similarity score. The final CLIP Score, derived from the mean of all frame scores, reflects the overall quality of textual alignment between the edited video and the target prompt.

\subsubsection{Auxiliary Metrics - Style and Consistency}
When considering the edited video independently of the original video and source prompt, it can be viewed as the output of a generative model. The last two evaluation dimensions, Style and Consistency, focus exclusively on the edited video itself. To complement our evaluation, we have carefully selected four metrics from VBench \cite{huang2024vbench}, originally designed for video generative models, to serve as auxiliary metrics for assessing video editing models. We have excluded most of the Video-Condition Consistency metrics, as the original video already defines the semantic structure and motion of the edited video. Furthermore, our video editing metrics are capable of assessing qualities such as temporal flickering and frame-wise consistency. Therefore, metrics for generative models serve as supplementary tools.

\noindent \textbf{Style - Aesthetic Quality.} 
Aesthetic Quality evaluates the artistic and aesthetic value perceived by humans towards each video frame using the LAION aesthetic predictor \cite{schuhmann2022laion}. Given the input of edited frames $(f_0', f_1',\ldots, f_n')$, the output is the average score $\frac{1}{n+1}\sum_{i=0}^{n} Q_1(f_i') $. This metric captures various aesthetic aspects, such as layout, color richness, photo-realism, naturalness, and overall artistic quality of edited frames.

\noindent \textbf{Style - Imaging Quality.}
Imaging Quality evaluates various types of distortion, such as over-exposure, noise, and blur, present in the edited frames $(f_0', f_1',\ldots, f_n')$ using the MUSIQ image quality predictor \cite{ke2021musiq} trained on the SPAQ dataset \cite{fang2020perceptual}. The output is the average score, calculated as $\frac{1}{n+1}\sum_{i=0}^{n} Q_2(f_i')$, offering a comprehensive assessment of the overall imaging quality of edited frames.

\noindent \textbf{Consistency - Subject Consistency.}
Subject Consistency measures the extent to which a subject’s appearance remains consistent across the entire video. Using DINO \cite{caron2021emerging} image features $(d_0, d_1, \ldots, d_n)$, where $d_i = DINO(f_i')$, we calculate the average score as $\frac{1}{n}\sum_{i=1}^{n} \frac{1}{2}(\langle d_0, d_i \rangle + \langle d_{i-1}, d_i \rangle) $. In this formula,  $\langle \cdot \rangle$ denotes the dot product operation for calculating cosine similarity. For each frame, the cosine similarity is computed with the first frame and its previous frame, and the average of these similarities is taken. The overall score is then derived by averaging across all non-starting video frames.

\noindent \textbf{Consistency - Background Consistency.}
Background Consistency evaluates the temporal consistency of the background scenes by calculating CLIP feature similarity across frames. Given the input of CLIP image features $(c_0, c_1, \ldots, c_n)$, where $c_i = CLIP(f_i')$, the output is the average score, calculated as $\frac{1}{n}\sum_{i=1}^{n} \frac{1}{2}(\langle c_0, c_i \rangle + \langle c_{i-1}, c_i \rangle) $. The calculation is similar to the method used for Subject Consistency. The only difference is that CLIP image features are used here instead of DINO image features.

\subsection{Task-Oriented Testing}
\begin{figure}[t]
\centering
\includegraphics[width=1\columnwidth]{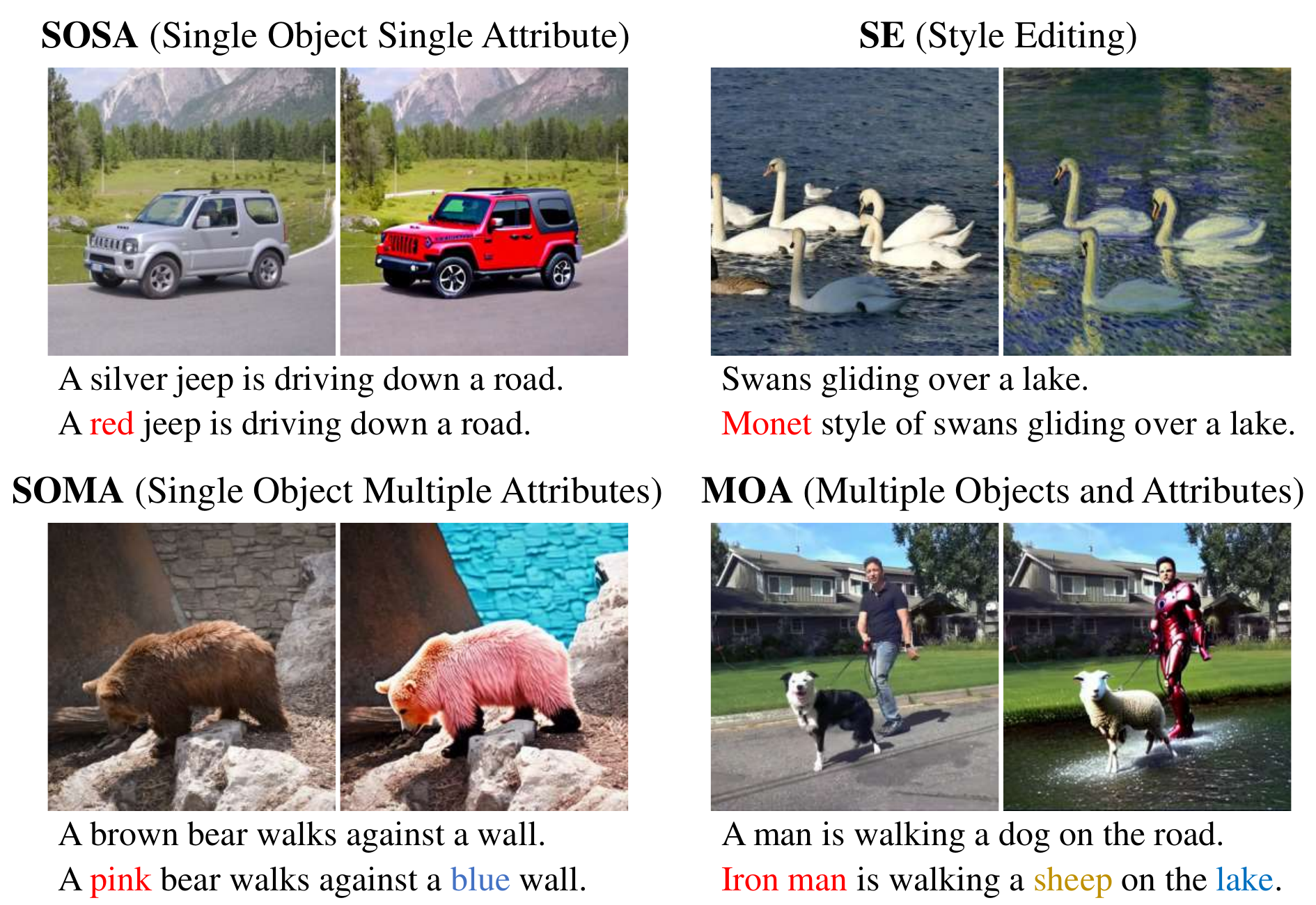} 
\caption{Categorization of video editing tasks.}
\label{fig3}
\end{figure}

To comprehensively evaluate each model, we define four main tasks in text-based video editing, as illustrated in Figure~\ref{fig3}. These tasks are further categorized into three complexity levels: simple, intermediate, and difficult. The simple level includes Single Object Single Attribute (SOSA) and Style Editing (SE), the intermediate level includes Single Object Multiple Attributes (SOMA), and the difficult level includes Multiple Objects and Attributes (MOA). The purpose of setting these levels is to break down model performance into details.

\noindent \textbf{Simple Level - Single Object Single Attribute (SOSA).}
The testing samples selected for SOSA tasks contain only one major object in the frame (e.g., a bear or a car), and the editing is performed solely on the object. Most models are capable of identifying the object and performing the edits. However, the real challenge is whether the model can accurately identify the object and leave the rest of the frames unchanged. Thus, Semantic Score is calculated through this task to assess how well the other parts of the edited frames align with the original ones.

\begin{table*}[htb]   
\begin{center}   
\resizebox{0.9\linewidth}{!}{
\begin{tabular}{c|c|c|c|c|c|c|c|c|c|c}
\Xhline{1pt}
& 
\textbf{Tasks}   & 
\textbf{FF-$\alpha$ $\downarrow $} & 
\textbf{FF-$\beta$ $\downarrow $} & \textbf{\Centerstack{Semantic\\Score $\downarrow $}} & \textbf{\Centerstack{Success\\Rate $\uparrow$}} & \textbf{\Centerstack{CLIP\\ Similarity $\uparrow$}} & \textbf{\Centerstack{Subject\\Consistency $\uparrow$}}   & \textbf{\Centerstack{Background\\Consistency $\uparrow$}} & \textbf{\Centerstack{Aesthetic\\Quality $\uparrow$}} & \textbf{\Centerstack{Imaging\\Quality $\uparrow$}}    \\
\Xhline{1pt}
\multirow{4}{*}{\centering FateZero} & SOSA & 8.0082 & 0.1723 & 25.2665 & 0.5294 &	0.3105 & 0.9696 & 0.9497 & 0.5546 & 0.6907\\ 
& SE  & 10.5420 & 0.5022 & - & 0.6923 &0.3341 & 0.9417 & 0.9563 & 0.5886 & 0.6267  \\
& SOMA & 10.2624 & 0.2795 & - & 0.6667 &	0.3108 & 0.9418 & 0.9280 & 0.5086 & 0.6083 \\
& MOA  & 12.2363 & 0.2832 & - & 0.4464 & 0.2944 & 0.9457 & 0.9449 & 0.4970 & 0.4867 \\

\hline
\hline

\multirow{4}{*}{\centering Control-A-Video} & SOSA & 18.0534 & 0.2674 & 66.1538 & 0.6050 &	0.3170 & 0.9672 & 0.9700 & 0.5377 & 0.6973\\ 
& SE  & 11.9252 & 0.5993 & - & 0.7143 &	0.3212 & 0.9440 & 0.9626 & 0.5575 & 0.7087 \\
& SOMA & 15.3962 & 0.3977 & - & 0.8429 &	0.3251 & 0.9595 & 0.9597 & 0.5751 & 0.7006 \\
& MOA  & 21.2607 & 0.5616 & - & 0.7347 &	0.3025 & 0.9249 & 0.9485 & 0.5085 & 0.7137 \\

\hline
\hline

\multirow{4}{*}{\centering Ground-A-Video} & SOSA & 6.0249 & 0.1022 & 8.2680 & 0.8659 &	0.3239 &	0.9622 &	0.9711 	&0.5635& 	0.6750 \\ 
& SE  & 7.7620 & 0.4259 & - & 0.9443 &	0.3452 &	0.9706 &	0.9503 	&0.5703 &	0.6914  \\
& SOMA & 7.3247 & 0.2293 & - & 0.8723 &	0.3479 &	0.9313 &	0.9486 	&0.5676 	&0.6673  \\
& MOA  & 7.2733 & 0.2337 & - & 0.8370 &	0.3340 &	0.9687 	&0.9502 &	0.5681 &	0.6737 \\

\hline
\hline

\multirow{4}{*}{\centering Video-P2P} & SOSA & 11.8893 & 0.2216 & 20.7714 & 0.5156 &	0.3037 &	0.9692 &	0.9696	&0.4847 & 	0.6665 \\ 
& SE  & 12.0104 & 0.5463 & - & 0.6058 &	0.3113 &	0.9561  &	0.9704 	&0.5125 &	0.5962  \\
& SOMA & 12.3277 & 0.3441 & - & 0.3462 &	0.2764 &	0.9657 &	0.9546 	&0.4827 	&0.6315  \\
& MOA  & 9.1412 & 0.4087 & - & 0.3752 &	0.2925 &	0.9533 	&0.9606 &	0.4921 &	0.5383 \\

\hline
\hline

\multirow{4}{*}{\centering TokenFlow} & SOSA & 7.2708 & 0.1566 & 31.6023 & 0.6471 &	0.3242 &	0.9790 &	0.9525 	&0.6233 & 	0.7408 \\ 
& SE  & 6.3735 & 0.4364 & - & 0.4135 &	0.3123 &	0.9762 &	0.9629 	&0.6421 &	0.7010  \\
& SOMA & 7.8735 & 0.2437  & - & 0.5750 &	0.3125 &	0.9739 &	0.9437 	&0.5842	& 0.6973  \\
& MOA  & 8.3205  & 0.3098  & - & 0.5532 &	0.3084 & 0.9546 &0.9595  &	0.5535 &	0.6759 \\

\Xhline{1pt}
\end{tabular}
}
% \caption{Transcript for FateZero. SOSA: Single Object Single Attribute; SE: Style Editing; SOMA: Single Object Multiple Attributes; MOA: Multiple Objects and Attributes.} 
\caption{Transcript for FateZero, Control-A-Video, Ground-A-Video, Video-P2P, and TokenFlow. SOSA: Single Object Single Attribute; SE: Style Editing; SOMA: Single Object Multiple Attributes; MOA: Multiple Objects and Attributes.}
\label{table1} 
\end{center}   
\end{table*}

\noindent \textbf{Simple Level - Style Editing (SE).}
Style Editing (SE) is a common task in video editing, involving changes to the global style of the video, such as converting the original video into a cyberpunk style. Based on our experience, most models perform well in this task. Therefore, we classify it as a simple level task.

\noindent \textbf{Intermediate Level - Single Object Multiple Attributes (SOMA).}
We further challenge the model on editing more than one attribute, requiring it to handle complex target prompts while maintaining consistency and fidelity. 

\noindent \textbf{Difficult Level - Multiple Objects and Attributes (MOA).}
The most challenging task involves editing multiple objects, requiring the model to be both precise and object-aware. Some models, such as EVA and Ground-A-Video, are specifically designed to address this challenge.

\begin{figure*}[t]
\centering
\includegraphics[width=2.1\columnwidth]{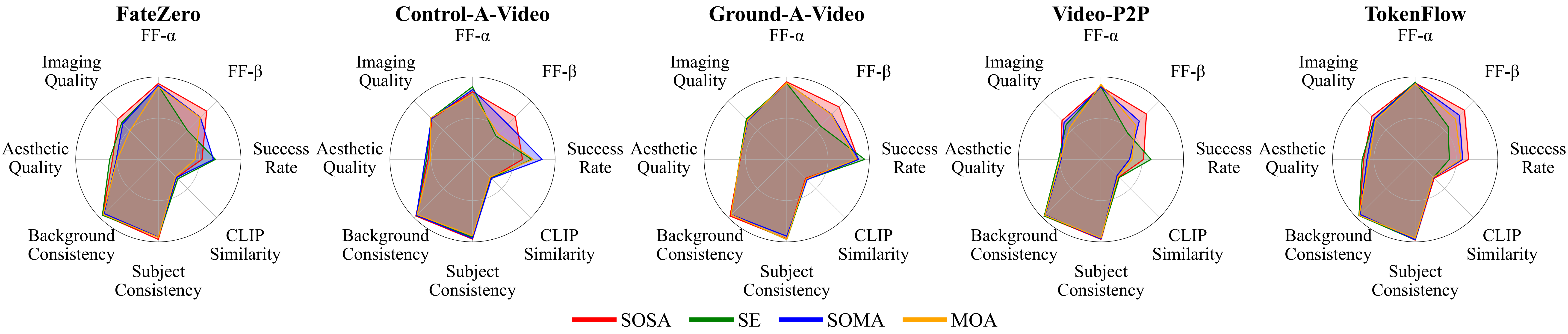} 
\caption{Visualization of FateZero, Control-A-Video, Ground-A-Video, Video-P2P, and TokenFlow's performance on four tasks. Most models perform worse at SOMA and MOA, verifying our categorization of tasks into different levels.}
\label{radar}
\end{figure*}

\subsection{Human Alignment}

We focus on metrics specifically designed for video editing models, as the metrics for generative models are already well-established. The objective of the human alignment experiment is to demonstrate that the automatic metrics presented in this paper align well with human perception. For the generation of testing data, we select three video editing models—FateZero, Control-A-Video, and TokenFlow—denoted as $M_1, M_2, M_3$, each provided with five source videos $S_1, S_2, S_3, S_4, S_5$ and corresponding target prompts (two for each source video, denoted as $P_i^1, P_i^2$). For example, source video $S_1$ is provided with target prompts $P_1^1$ and $P_1^2$. We conduct parallel comparisons, meaning the comparison occurs only between edited videos from the same source video and prompt. This setup results in ten groups of edited videos, each containing three outputs from the three models. For example:
$G_1 = \left\{M1(S_1, P_1^1), M2(S_1, P_1^1), M3(S_1, P_1^1)\right\}$
For each group, the videos are paired up, yielding $C_{3}^2 = 3$ comparisons so that human annotators can make direct comparisons between only two choices at a time. For each evaluation dimension, we instruct the human annotators to consider only the specific aspect being evaluated.
We engage 30 individuals to annotate their preferences. They are asked to select the video that performs better on the specified evaluation dimension. Further details regarding the human annotation experiment can be found in the \textbf{Experiment} section.
\section{Experiment}

This section presents the evaluation experiment conducted using EditBoard, along with the human alignment test, which is specifically designed to validate the correlation between automatic metrics and human perception. We rigorously evaluate five state-of-the-art video editing models: FateZero \cite{qi2023fatezero}, Control-A-Video \cite{chen2023control}, Ground-A-Video \cite{jeong2023ground}, TokenFlow \cite{geyer2023tokenflow}, and Video-P2P \cite{liu2024video}. For each model, EditBoard generates a detailed transcript that reports performance across each dimension and task type. To ensure a fair comparison, we use Stable Diffusion v1-5 as the base for all video editing models. The experiments are conducted on a single NVIDIA GeForce RTX™ 4090.

\subsection{Data Preparation}
We utilize samples from the DAVIS dataset \cite{pont20172017} to obtain the masks required for conducting Semantic Score testing. We also select samples from the LOVEU-TGVE-2023 dataset \cite{wu2023cvpr}. For each task, we select 10 videos containing a variety of objects, such as cars, animals, and humans. Each original video is paired with at least two target prompts according to the task category. For generating source prompts, we employ BLIP-2 \cite{li2023blip} for the automated generation of video captions. The original frames are resized to a uniform resolution of 512$\times$512 to match the configuration of the testing models. We also ensure that sufficient original videos meet the requirements for applying FF-$\alpha$, allowing for the full adoption of both FF-$\alpha$ and FF-$\beta$. Additionally, we adjust the target prompts for Single Object Single Attribute (SOSA) so that more than half of the edits focus on the foreground object to facilitate Semantic Score evaluation.

\subsection{Task-Oriented Evaluation}
A transcript is produced for each model, detailing the scores attained per metric for each task, as shown in Table~\ref{table1}. The visualization of each model's performance is shown in Figure~\ref{radar}. We will include more models as they become open-sourced. The results indicate that model performance tends to decline as the task difficulty increases from simple to difficult, supporting the validity of our initial categorization.

\subsection{Human Alignment Experiment}
We conduct a human alignment experiment to evaluate whether the automatic metrics align with human perception. This is done by calculating the percentage of questions where human judgments match the outcomes suggested by the metrics, as shown in Table~\ref{table2}. In this experiment, human annotators compare two samples based on a specific evaluation criterion and select the one with higher quality or indicate that the difference is negligible. A match is recorded if the human choice agrees with the metrics’ comparison. Also, if annotators mark the samples as “hard to distinguish” and the metrics’ values differ by less than a predefined threshold, it is also considered a match.

\begin{table}[htb]   
\begin{center}   
\resizebox{\linewidth}{!}{
\begin{tabular}{c|c|c|c|c|c}
\Xhline{1pt}
    &FF-$\alpha$ & FF-$\beta$ & \Centerstack{Semantic\\Score} & \Centerstack{Success\\Rate} & \Centerstack{CLIP\\ Similarity} \\
\hline
\textbf{Matching Rate} & 92.54 & 89.93 & 95.24 & 92.68 & 85.72 \\ 

\Xhline{1pt}
\end{tabular}
}
\caption{Our new metrics achieve around 90\% match with human choices.}  
\label{table2} 
\end{center}   
\end{table}

\noindent We also provide the Pearson's correlation score in Table~\ref{table3} to further validate that our new metrics align with human perception.

\begin{table}[htb]   
\begin{center}   
\resizebox{\linewidth}{!}{
\begin{tabular}{c|c|c|c|c|c}
\Xhline{1pt}
    &FF-$\alpha$ & FF-$\beta$ & \Centerstack{Semantic\\Score} & \Centerstack{Success\\Rate} & \Centerstack{CLIP\\ Similarity} \\
\hline
\textbf{Pearson's score}& 0.9312 & 0.9126 & 	0.9417 &	0.9333 & 0.8769\\ 

\Xhline{1pt}
\end{tabular}
}
\caption{Pearson correlation scores demonstrate a similar trend, reinforcing the alignment.}  
\label{table3} 
\end{center}   
\end{table}

\noindent Additionally, we evaluate FF-$\alpha$ and FF-$\beta$ for capturing temporal flickering and motion consistency, finding that higher FF scores (lower quality) align with more pronounced issues. This suggests traditional metrics like motion smoothness may be unnecessary for video editing evaluation.

\section{Insights and Discussions}

\subsection{A Transcript as a Diagnostic Tool}

The transcript provides comprehensive insights into the strengths and weaknesses of the models. By analyzing the transcript, we can quickly identify the model's limitations and explore potential causes. For instance, when comparing the transcript of FateZero with that of Control-A-Video, we observe that FateZero shows significant improvement in Semantic Score, indicating better object-aware editing. However, it shows little improvement in Success Rate or CLIP Similarity for the SOSA task. This contrast highlights FateZero's deficiency in editing individual objects (such as turning a rabbit into a squirrel). Furthermore, FateZero's higher Semantic Score compared to Ground-A-Video suggests a deeper issue with attention blending. Specifically, given that FateZero adopts an attention blending method, the unedited parts should remain mostly unchanged—yet they do not. This inconsistency likely stems from inaccurate attention, causing unintended parts to be edited. For example, when turning a silver jeep into a red jeep, the road also turns red. The attention leakage problem is also corroborated by the EVA paper \cite{yang2024eva}.

\subsection{Trade-off Between Execution and Fidelity}

Our experiments reveal an intriguing trade-off between execution and fidelity. Models scoring higher in execution tend to perform worse in fidelity. For example, FateZero achieves better scores in FF-$\alpha$, FF-$\beta$, and Semantic Score across all tasks but underperforms in Success Rate and CLIP Similarity. This discrepancy can be attributed to their respective methodologies. Control-A-Video manipulates latent space during the diffusion process, resulting in more structural changes and better adherence to target prompts. In contrast, FateZero's attention blending approach is more conservative, preserving the original structure but compromising execution. Thus, the model's editing method plays a crucial role in balancing execution and fidelity.

% \subsection{Style Editing Helps Execution}

% An interesting observation from our experiments is that some models tend to achieve higher execution scores on SOMA tasks than SOSA tasks, despite the former's increased complexity. This anomaly is partly due to some SOMA target prompts involving editing multiple attributes along with the global style. Upon reviewing the results, we find that models execute object editing more effectively when the style becomes more abstract. For example, FateZero struggles to transform a man into an ape in SOMA. However, when an additional prompt to apply a Van Gogh style is added, the execution improves significantly (see Figure~\ref{fig8}). Future research can leverage this finding to explore how style editing can enhance execution, potentially offering new avenues for improving model performance. More discussions are provided in Appendix Section 7.

% \begin{figure}[t]
% \centering
% \includegraphics[width=\columnwidth]{ape3.pdf} 
% \caption{Despite the structural change in Edit Result 2, the model successfully turns the man into an ape with the additional prompt of applying Van Gogh painting style.}
% \label{fig8}
% \end{figure}

\section{Conclusion}

In this paper, we propose EditBoard, a pioneering comprehensive evaluation benchmark specifically designed for text-based video editing models. Our benchmark addresses the critical need for a standardized framework that holistically assesses the multifaceted performance of these models. By incorporating nine metrics across four dimensions and introducing three novel metrics, EditBoard provides a detailed, task-oriented evaluation that highlights each model's strengths and weaknesses. Empirical results demonstrate EditBoard's efficacy in aligning with human perceptions of video quality and editing precision. By open-sourcing EditBoard, we aim to foster the development of more robust and reliable video editing models, ultimately advancing the field of AIGC. Our work sets a new standard for evaluating text-based video editing models, ensuring a more comprehensive and objective assessment for future research.

\clearpage
\section*{Acknowledgments}
We sincerely thank the reviewers for their constructive feedback and insightful suggestions, which have greatly improved this work. We also extend our heartfelt gratitude to Prof. Baoyuan Wu, Prof. Mingqiang Wei, Prof. Limin Wang, and Dr. Peng Zhou for their invaluable guidance and support throughout this research.

\bibliography{aaai25}

\begin{thebibliography}{39}
\providecommand{\natexlab}[1]{#1}

\bibitem[{Blattmann et~al.(2023)Blattmann, Rombach, Ling, Dockhorn, Kim, Fidler, and Kreis}]{blattmann2023align}
Blattmann, A.; Rombach, R.; Ling, H.; Dockhorn, T.; Kim, S.~W.; Fidler, S.; and Kreis, K. 2023.
\newblock Align your latents: High-resolution video synthesis with latent diffusion models.
\newblock In \emph{Proceedings of the IEEE/CVF Conference on Computer Vision and Pattern Recognition}, 22563--22575.

\bibitem[{Caron et~al.(2021)Caron, Touvron, Misra, J{\'e}gou, Mairal, Bojanowski, and Joulin}]{caron2021emerging}
Caron, M.; Touvron, H.; Misra, I.; J{\'e}gou, H.; Mairal, J.; Bojanowski, P.; and Joulin, A. 2021.
\newblock Emerging properties in self-supervised vision transformers.
\newblock In \emph{Proceedings of the IEEE/CVF international conference on computer vision}, 9650--9660.

\bibitem[{Chai et~al.(2023)Chai, Guo, Wang, and Lu}]{chai2023stablevideo}
Chai, W.; Guo, X.; Wang, G.; and Lu, Y. 2023.
\newblock Stablevideo: Text-driven consistency-aware diffusion video editing.
\newblock In \emph{Proceedings of the IEEE/CVF International Conference on Computer Vision}, 23040--23050.

\bibitem[{Chang, Chen, and Liu(2023)}]{chang2023diffusionatlas}
Chang, S.-Y.; Chen, H.-T.; and Liu, T.-L. 2023.
\newblock DiffusionAtlas: High-Fidelity Consistent Diffusion Video Editing.
\newblock \emph{arXiv preprint arXiv:2312.03772}.

\bibitem[{Chen et~al.(2023{\natexlab{a}})Chen, Xia, He, Zhang, Cun, Yang, Xing, Liu, Chen, Wang et~al.}]{chen2023videocrafter1}
Chen, H.; Xia, M.; He, Y.; Zhang, Y.; Cun, X.; Yang, S.; Xing, J.; Liu, Y.; Chen, Q.; Wang, X.; et~al. 2023{\natexlab{a}}.
\newblock Videocrafter1: Open diffusion models for high-quality video generation.
\newblock \emph{arXiv preprint arXiv:2310.19512}.

\bibitem[{Chen et~al.(2023{\natexlab{b}})Chen, Ji, Wu, Wu, Xie, Li, Xia, Xiao, and Lin}]{chen2023control}
Chen, W.; Ji, Y.; Wu, J.; Wu, H.; Xie, P.; Li, J.; Xia, X.; Xiao, X.; and Lin, L. 2023{\natexlab{b}}.
\newblock Control-a-video: Controllable text-to-video generation with diffusion models.
\newblock \emph{arXiv preprint arXiv:2305.13840}.

\bibitem[{Fang et~al.(2020)Fang, Zhu, Zeng, Ma, and Wang}]{fang2020perceptual}
Fang, Y.; Zhu, H.; Zeng, Y.; Ma, K.; and Wang, Z. 2020.
\newblock Perceptual quality assessment of smartphone photography.
\newblock In \emph{Proceedings of the IEEE/CVF conference on computer vision and pattern recognition}, 3677--3686.

\bibitem[{Ge et~al.(2023)Ge, Nah, Liu, Poon, Tao, Catanzaro, Jacobs, Huang, Liu, and Balaji}]{ge2023preserve}
Ge, S.; Nah, S.; Liu, G.; Poon, T.; Tao, A.; Catanzaro, B.; Jacobs, D.; Huang, J.-B.; Liu, M.-Y.; and Balaji, Y. 2023.
\newblock Preserve your own correlation: A noise prior for video diffusion models.
\newblock In \emph{Proceedings of the IEEE/CVF International Conference on Computer Vision}, 22930--22941.

\bibitem[{Geyer et~al.(2023)Geyer, Bar-Tal, Bagon, and Dekel}]{geyer2023tokenflow}
Geyer, M.; Bar-Tal, O.; Bagon, S.; and Dekel, T. 2023.
\newblock TokenFlow: Consistent Diffusion Features for Consistent Video Editing.
\newblock In \emph{The Twelfth International Conference on Learning Representations}.

\bibitem[{Guo et~al.(2023)Guo, Yang, Rao, Liang, Wang, Qiao, Agrawala, Lin, and Dai}]{guo2023animatediff}
Guo, Y.; Yang, C.; Rao, A.; Liang, Z.; Wang, Y.; Qiao, Y.; Agrawala, M.; Lin, D.; and Dai, B. 2023.
\newblock AnimateDiff: Animate Your Personalized Text-to-Image Diffusion Models without Specific Tuning.
\newblock In \emph{The Twelfth International Conference on Learning Representations}.

\bibitem[{Gupta et~al.(2023)Gupta, Yu, Sohn, Gu, Hahn, Fei-Fei, Essa, Jiang, and Lezama}]{gupta2023photorealistic}
Gupta, A.; Yu, L.; Sohn, K.; Gu, X.; Hahn, M.; Fei-Fei, L.; Essa, I.; Jiang, L.; and Lezama, J. 2023.
\newblock Photorealistic video generation with diffusion models.
\newblock \emph{arXiv preprint arXiv:2312.06662}.

\bibitem[{He et~al.(2022)He, Yang, Zhang, Shan, and Chen}]{he2022latent}
He, Y.; Yang, T.; Zhang, Y.; Shan, Y.; and Chen, Q. 2022.
\newblock Latent video diffusion models for high-fidelity video generation with arbitrary lengths.
\newblock \emph{arXiv preprint arXiv:2211.13221}, 2(3): 4.

\bibitem[{Hessel et~al.(2021)Hessel, Holtzman, Forbes, Le~Bras, and Choi}]{hessel2021clipscore}
Hessel, J.; Holtzman, A.; Forbes, M.; Le~Bras, R.; and Choi, Y. 2021.
\newblock CLIPScore: A Reference-free Evaluation Metric for Image Captioning.
\newblock In \emph{Proceedings of the 2021 Conference on Empirical Methods in Natural Language Processing}, 7514--7528.

\bibitem[{Ho, Jain, and Abbeel(2020)}]{ho2020denoising}
Ho, J.; Jain, A.; and Abbeel, P. 2020.
\newblock Denoising diffusion probabilistic models.
\newblock \emph{Advances in neural information processing systems}, 33: 6840--6851.

\bibitem[{Hu and Xu(2023)}]{hu2023videocontrolnet}
Hu, Z.; and Xu, D. 2023.
\newblock Videocontrolnet: A motion-guided video-to-video translation framework by using diffusion model with controlnet.
\newblock \emph{arXiv preprint arXiv:2307.14073}.

\bibitem[{Huang et~al.(2024)Huang, He, Yu, Zhang, Si, Jiang, Zhang, Wu, Jin, Chanpaisit et~al.}]{huang2024vbench}
Huang, Z.; He, Y.; Yu, J.; Zhang, F.; Si, C.; Jiang, Y.; Zhang, Y.; Wu, T.; Jin, Q.; Chanpaisit, N.; et~al. 2024.
\newblock Vbench: Comprehensive benchmark suite for video generative models.
\newblock In \emph{Proceedings of the IEEE/CVF Conference on Computer Vision and Pattern Recognition}, 21807--21818.

\bibitem[{Ilg et~al.(2017)Ilg, Mayer, Saikia, Keuper, Dosovitskiy, and Brox}]{ilg2017flownet}
Ilg, E.; Mayer, N.; Saikia, T.; Keuper, M.; Dosovitskiy, A.; and Brox, T. 2017.
\newblock Flownet 2.0: Evolution of optical flow estimation with deep networks.
\newblock In \emph{Proceedings of the IEEE conference on computer vision and pattern recognition}, 2462--2470.

\bibitem[{Jeong and Ye(2023)}]{jeong2023ground}
Jeong, H.; and Ye, J.~C. 2023.
\newblock Ground-a-video: Zero-shot grounded video editing using text-to-image diffusion models.
\newblock \emph{arXiv preprint arXiv:2310.01107}.

\bibitem[{Ke et~al.(2021)Ke, Wang, Wang, Milanfar, and Yang}]{ke2021musiq}
Ke, J.; Wang, Q.; Wang, Y.; Milanfar, P.; and Yang, F. 2021.
\newblock Musiq: Multi-scale image quality transformer.
\newblock In \emph{Proceedings of the IEEE/CVF international conference on computer vision}, 5148--5157.

\bibitem[{Ku et~al.(2024)Ku, Wei, Ren, Yang, and Chen}]{ku2024anyv2vtuningfreeframeworkvideotovideo}
Ku, M.; Wei, C.; Ren, W.; Yang, H.; and Chen, W. 2024.
\newblock AnyV2V: A Tuning-Free Framework For Any Video-to-Video Editing Tasks.
\newblock arXiv:2403.14468.

\bibitem[{Li et~al.(2019)Li, Qi, Lukasiewicz, and Torr}]{li2019controllable}
Li, B.; Qi, X.; Lukasiewicz, T.; and Torr, P. 2019.
\newblock Controllable text-to-image generation.
\newblock \emph{Advances in neural information processing systems}, 32.

\bibitem[{Li et~al.(2023)Li, Li, Savarese, and Hoi}]{li2023blip}
Li, J.; Li, D.; Savarese, S.; and Hoi, S. 2023.
\newblock Blip-2: Bootstrapping language-image pre-training with frozen image encoders and large language models.
\newblock In \emph{International conference on machine learning}, 19730--19742. PMLR.

\bibitem[{Liu et~al.(2024)Liu, Zhang, Li, Lin, and Jia}]{liu2024video}
Liu, S.; Zhang, Y.; Li, W.; Lin, Z.; and Jia, J. 2024.
\newblock Video-p2p: Video editing with cross-attention control.
\newblock In \emph{Proceedings of the IEEE/CVF Conference on Computer Vision and Pattern Recognition}, 8599--8608.

\bibitem[{Luo et~al.(2023)Luo, Chen, Zhang, Huang, Wang, Shen, Zhao, Zhou, and Tan}]{luo2023videofusion}
Luo, Z.; Chen, D.; Zhang, Y.; Huang, Y.; Wang, L.; Shen, Y.; Zhao, D.; Zhou, J.; and Tan, T. 2023.
\newblock Videofusion: Decomposed diffusion models for high-quality video generation.
\newblock In \emph{2023 IEEE/CVF Conference on Computer Vision and Pattern Recognition (CVPR)}, 10209--10218. IEEE.

\bibitem[{Nichol et~al.(2022)Nichol, Dhariwal, Ramesh, Shyam, Mishkin, Mcgrew, Sutskever, and Chen}]{nichol2022glide}
Nichol, A.~Q.; Dhariwal, P.; Ramesh, A.; Shyam, P.; Mishkin, P.; Mcgrew, B.; Sutskever, I.; and Chen, M. 2022.
\newblock GLIDE: Towards Photorealistic Image Generation and Editing with Text-Guided Diffusion Models.
\newblock In \emph{International Conference on Machine Learning}, 16784--16804. PMLR.

\bibitem[{Parmar et~al.(2023)Parmar, Kumar~Singh, Zhang, Li, Lu, and Zhu}]{parmar2023zero}
Parmar, G.; Kumar~Singh, K.; Zhang, R.; Li, Y.; Lu, J.; and Zhu, J.-Y. 2023.
\newblock Zero-shot image-to-image translation.
\newblock In \emph{ACM SIGGRAPH 2023 Conference Proceedings}, 1--11.

\bibitem[{Pont-Tuset et~al.(2017)Pont-Tuset, Perazzi, Caelles, Arbel{\'a}ez, Sorkine-Hornung, and Van~Gool}]{pont20172017}
Pont-Tuset, J.; Perazzi, F.; Caelles, S.; Arbel{\'a}ez, P.; Sorkine-Hornung, A.; and Van~Gool, L. 2017.
\newblock The 2017 davis challenge on video object segmentation.
\newblock \emph{arXiv preprint arXiv:1704.00675}.

\bibitem[{Qi et~al.(2023)Qi, Cun, Zhang, Lei, Wang, Shan, and Chen}]{qi2023fatezero}
Qi, C.; Cun, X.; Zhang, Y.; Lei, C.; Wang, X.; Shan, Y.; and Chen, Q. 2023.
\newblock Fatezero: Fusing attentions for zero-shot text-based video editing.
\newblock In \emph{Proceedings of the IEEE/CVF International Conference on Computer Vision}, 15932--15942.

\bibitem[{Radford et~al.(2021)Radford, Kim, Hallacy, Ramesh, Goh, Agarwal, Sastry, Askell, Mishkin, Clark et~al.}]{radford2021learning}
Radford, A.; Kim, J.~W.; Hallacy, C.; Ramesh, A.; Goh, G.; Agarwal, S.; Sastry, G.; Askell, A.; Mishkin, P.; Clark, J.; et~al. 2021.
\newblock Learning transferable visual models from natural language supervision.
\newblock In \emph{International conference on machine learning}, 8748--8763. PMLR.

\bibitem[{Rombach et~al.(2022)Rombach, Blattmann, Lorenz, Esser, and Ommer}]{rombach2022high}
Rombach, R.; Blattmann, A.; Lorenz, D.; Esser, P.; and Ommer, B. 2022.
\newblock High-resolution image synthesis with latent diffusion models.
\newblock In \emph{Proceedings of the IEEE/CVF conference on computer vision and pattern recognition}, 10684--10695.

\bibitem[{Schuhmann(2022)}]{schuhmann2022laion}
Schuhmann, C. 2022.
\newblock LAION Aesthetic Predictor.
\newblock \url{https://laion.ai/blog/laion-aesthetics/}.

\bibitem[{Sohl-Dickstein et~al.(2015)Sohl-Dickstein, Weiss, Maheswaranathan, and Ganguli}]{sohl2015deep}
Sohl-Dickstein, J.; Weiss, E.; Maheswaranathan, N.; and Ganguli, S. 2015.
\newblock Deep unsupervised learning using nonequilibrium thermodynamics.
\newblock In \emph{International conference on machine learning}, 2256--2265. PMLR.

\bibitem[{Sun et~al.(2018)Sun, Yang, Liu, and Kautz}]{sun2018pwc}
Sun, D.; Yang, X.; Liu, M.-Y.; and Kautz, J. 2018.
\newblock Pwc-net: Cnns for optical flow using pyramid, warping, and cost volume.
\newblock In \emph{Proceedings of the IEEE conference on computer vision and pattern recognition}, 8934--8943.

\bibitem[{Sun et~al.(2024)Sun, Tu, Liao, and Tao}]{sun2024diffusion}
Sun, W.; Tu, R.-C.; Liao, J.; and Tao, D. 2024.
\newblock Diffusion Model-Based Video Editing: A Survey.
\newblock \emph{arXiv preprint arXiv:2407.07111}.

\bibitem[{Villegas et~al.(2022)Villegas, Babaeizadeh, Kindermans, Moraldo, Zhang, Saffar, Castro, Kunze, and Erhan}]{villegas2022phenaki}
Villegas, R.; Babaeizadeh, M.; Kindermans, P.-J.; Moraldo, H.; Zhang, H.; Saffar, M.~T.; Castro, S.; Kunze, J.; and Erhan, D. 2022.
\newblock Phenaki: Variable length video generation from open domain textual descriptions.
\newblock In \emph{International Conference on Learning Representations}.

\bibitem[{Wu et~al.(2023{\natexlab{a}})Wu, Ge, Wang, Lei, Gu, Shi, Hsu, Shan, Qie, and Shou}]{wu2023tune}
Wu, J.~Z.; Ge, Y.; Wang, X.; Lei, S.~W.; Gu, Y.; Shi, Y.; Hsu, W.; Shan, Y.; Qie, X.; and Shou, M.~Z. 2023{\natexlab{a}}.
\newblock Tune-a-video: One-shot tuning of image diffusion models for text-to-video generation.
\newblock In \emph{Proceedings of the IEEE/CVF International Conference on Computer Vision}, 7623--7633.

\bibitem[{Wu et~al.(2023{\natexlab{b}})Wu, Li, Gao, Dong, Bai, Singh, Xiang, Li, Huang, Sun, He, Hu, Hu, Huang, Zhu, Cheng, Tang, Shou, Keutzer, and Iandola}]{wu2023cvpr}
Wu, J.~Z.; Li, X.; Gao, D.; Dong, Z.; Bai, J.; Singh, A.; Xiang, X.; Li, Y.; Huang, Z.; Sun, Y.; He, R.; Hu, F.; Hu, J.; Huang, H.; Zhu, H.; Cheng, X.; Tang, J.; Shou, M.~Z.; Keutzer, K.; and Iandola, F. 2023{\natexlab{b}}.
\newblock CVPR 2023 Text Guided Video Editing Competition.
\newblock arXiv:2310.16003.

\bibitem[{Yang et~al.(2024)Yang, Zhu, Fan, and Yang}]{yang2024eva}
Yang, X.; Zhu, L.; Fan, H.; and Yang, Y. 2024.
\newblock EVA: Zero-shot Accurate Attributes and Multi-Object Video Editing.
\newblock \emph{arXiv preprint arXiv:2403.16111}.

\bibitem[{Zhang et~al.(2018)Zhang, Isola, Efros, Shechtman, and Wang}]{zhang2018unreasonable}
Zhang, R.; Isola, P.; Efros, A.~A.; Shechtman, E.; and Wang, O. 2018.
\newblock The unreasonable effectiveness of deep features as a perceptual metric.
\newblock In \emph{Proceedings of the IEEE conference on computer vision and pattern recognition}, 586--595.

\end{thebibliography}

\end{document}